# Comparative Study of Predicting Stock Index Using Deep Learning Models

Harshal Patel[1], Bharath Kumar Bolla[2], Sabeesh E[3], Dinesh Reddy[4]

[1]Liverpool John Moores University, London
 sabeesh90@yahoo.co.uk

[2]Salesforce, Hyderabad, India
 bolla111@gmail.com

[3]Liverpool John Moores University, London
 sabeesh90@yahoo.co.uk

[4]Cardinal Health, Bengaluru, India
 dinereddy1717@gmail.com

**Abstract.** Time series forecasting has seen many methods attempted over the past few decades, including traditional technical analysis, algorithmic statistical models, and more recent machine learning and artificial intelligence approaches. Recently, neural networks have been incorporated into the forecasting scenario, such as the LSTM and conventional RNN approaches, which utilize short-term and long-term dependencies. This study evaluates traditional forecasting methods, such as ARIMA, SARIMA, and SARIMAX, and newer neural network approaches, such as DF-RNN, DSSM, and Deep AR, built using RNNs. The standard NIFTY-50 dataset from Kaggle is used to assess these models using metrics such as MSE, RMSE, MAPE, POCID, and Theil's U. Results show that Deep AR outperformed all other conventional deep learning and traditional approaches, with the lowest MAPE of 0.01 and RMSE of 189. Additionally, the performance of Deep AR and GRU did not degrade when the amount of training data was reduced, suggesting that these models may not require a large amount of data to achieve consistent and reliable performance. The study demonstrates that incorporating deep learning approaches in a forecasting scenario significantly outperforms conventional approaches and can handle complex datasets, with potential applications in various domains, such as weather predictions and other time series applications in a real-world scenario.

Time series forecasting has been implemented traditionally using standard methods such as ARIMA, SARIMA, and SARIMAX [1]. A significant drawback of these methods has been their inability to handle multivariate datasets where exogenous variables significantly affect the forecasting predictions [2]. Furthermore, their accuracies of predictions have not been satisfying enough in many complex real-world scenarios [2]. The advent of Deep Learning has helped bridge this gap. Neural networks and their ability to achieve universal approximation is a well-established theory, as seen in scenarios such as regression and classification [3]. In the last two decades, models based on recurrent neural networks (RNNs) and LSTMs (Long-short-term memory) have been widely used in the forecasting scenario with promising results, and their ability to process sequential data has been exploited to solve complex time series scenarios [4]. However, in the last decade, newer architectures such as Deep Factor RNN(DF-RNN) [5], DSSM [6], Deep AR [7], and Deep Renewal [8] have been shown to outperform classical RNN and LSTM-based deep learning models in various scenarios. Very little experimentation has been done using these approaches on the Stock Market data. Hence, a comparative study of these models on a widely established dataset such as the NIFTY 50 index would help establish the superiority of deep learning models over traditional approaches and evaluate the effectiveness of recent deep learning models. The research objectives are as follows.

- To evaluate the superiority of neural networks over traditional approaches in forecasting augmentation.
- To evaluate the performance of the models, varying levels of train data (50% and 25%), keeping the test data the same to assess the effect of lesser data on model's performance.
- To evaluate if the better models are performing consistently on all metrics (MSE, RMSE, POCID, Thelis'U, MAPE.

## 1 Literature Review

Various research has been done in the recent past to increase the efficiency of time series forecasting by incorporating deep learning methodologies. While traditional approaches have been used to solve time series problems in a univariate scenario, deep learning approaches have been used to approximate multivariate datasets with significantly higher efficiency.

### 1.1 Traditional Approaches

Time series forecasting has been an important research field since humans started to predict values associated with a time component. According to De Gooijer and Hyndman [9], the earliest statistical models for time series analysis, namely the Auto Regressive (AR) and Moving Average (MA) models, were developed in the 1940s.

These models aimed to describe time series autocorrelation and were limited to linear forecasting problems. As researchers delved deeper into the subject, they factored in parametric influences. In the early 1970s, Box et al. [10] developed the Box-Jenkins method, a three-step iterative process for determining time series, which became a popular approach for time-series modeling. With the advent of computers and increasing processing power, Autoregressive integrated Moving Average (ARIMA) models were used empirically for univariate and multivariate time series forecasting. In the 1980s and 1990s, researchers started incorporating seasonality in time series modeling. Various methods, including X-11, X-12-ARIMA, etc., used decomposition to obtain seasonality and apply it in time-series forecasting [11].

In the past few decades, many methods have been implemented to forecast various domains. These methods include simple traditional technical analysis (also known as "charting") of price charts [12], algorithmic statistical models [13], and more recent Machine Learning and Artificial Intelligent approaches [14]. Computational time series forecasting has applications in various fields, from weather and sales forecasting to finance-related forecasting (budget analysis, stock market price forecasting). It is an indispensable tool for all fields that rely on time factors. Methods including Autoregression, Box Jenkins, and Holt-Winters were used to yield generally acceptable results.

### 1.2 Deep Learning Approaches

Recently, novel techniques and models have emerged utilizing deep learning methodologies. For instance, the Long- and Short-Term Time-Series Network (LSTNet) incorporates Convolutional Neural Network (CNN) and Recurrent Neural Network (RNN) to capture both short-term and long-term dependencies in time-series data [15]. Another approach proposes using the Gaussian Copula process (GP-Copula) in conjunction with RNN [16]. The Neural Basis Expansion Analysis for Interpretable Time Series forecasting (NBEATS) achieved state-of-the-art performance in the recent M4 time-series prediction competition [17]. It has been observed that deep learning methods possess an edge over traditional techniques with regard to overfitting, as evidenced in previous research by [18]

Many studies have shown that classical deep learning and machine learning models outperform ARIMA models in time-series forecasting. Various complex models, including Multi-Layer Perceptron, CNN, and LSTM, have been implemented and analyzed for time-series forecasting. These models can handle multiple input features, leading to higher accuracy than conventional methods. Feature extraction is a critical step in improving the performance of predictive models, even when simple features are used. Some studies have used modified deep networks to extract frequency-related features from time-series data using EMD and Complete Ensemble Empirical Mode Decomposition with Adaptive Noise (CEEDMAN). The extracted features were then fed to LSTM to predict one-step-ahead forecasting [19], [20]. Other studies have used image data features by decomposing raw time-series data into IMFs using IF and providing CNN to learn features automatically [21]. Data augmentation approaches such as adding external text-based sentiment data to the model-generated features, were also used [22]

Additionally, auto-regressive models have been proposed, such as DeepAR , which uses high-dimensional related time-series features to train Autoregressive Recurrent Neural Networks and has demonstrated superior performance compared to other competitive models [7]. Another study proposed a Multi-Step Time-Series Forecaster that uses various related time-series features to forecast demand on Amazon.com [23]. Furthermore, several state-of-the-art methods have been developed and proven to be highly promising in generalized competitions like M4 [24]. Finally, it has been shown that an ensemble of models consistently performs better than any single model [25].

## 2 Research Methodology

A novel python-based library, GluonTS, has been introduced to provide models, tools, and components for time-series forecasting [26]. Techniques such as DF-RNN, DSSM, Deep AR, and LSTNet have been implemented using relevant libraries from the GluonTS framework. The models used as baselines are ARIMA, SARIMA, SARIMAX, and Facebook's Prophet. Facebook's Prophet API is used for implementing the Prophet model.

### 2.1 Dataset - Exploration

The dataset used in this research consists of the NIFTY 50 index consisting of the closing value of the stock indices from Jan 2011 till Feb 2022 (Figure 1). A rolling window of length ten has been taken with context and prediction length of five each. Mean, and Standard deviation is calculated for ten window period to test for the stationarity of the time series. As seen in Fig 2, where there is a variation in the mean and Standard deviation across the ten-window time frame, it is evident that the time series is not stationary. This is further confirmed by the ADF test, as seen in Table 1. Indices from Jan 2011 to 2020 have been used as Train, and the subsequent series have been used as Test data, as seen in Figure 1.

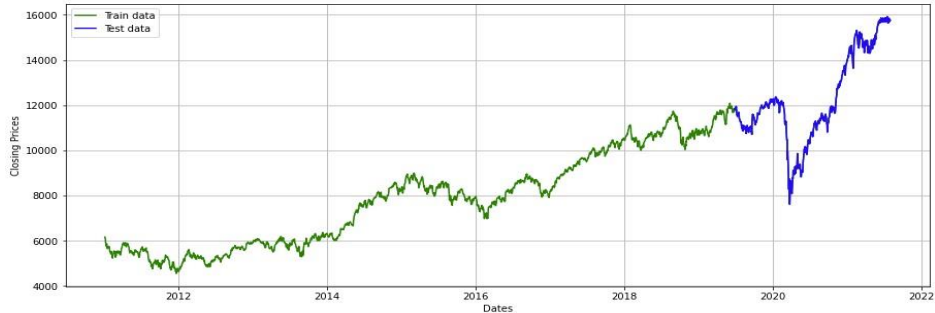
*Figure 1. Train test split – NIFTY 50 index*

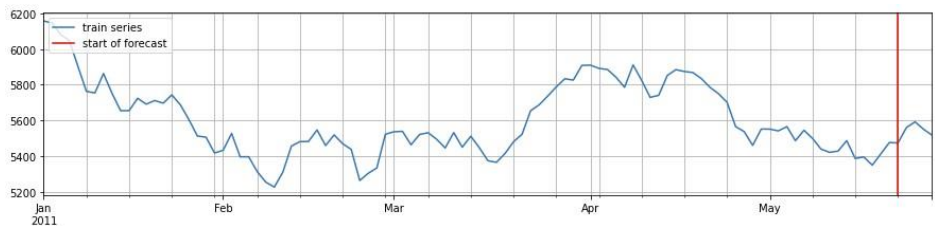
*Figure 2. Ten period Rolling window – Non uniform mean and STD – Non-stationary series*

**2.2 Forecasting Models**

Various models used in the experiments have been elaborated in the succeeding sections. Baseline forecasting models have been built using machine learning models such as ARIMA, SARIMA, and SARIMAX, while deep learning models have been built using DF-RNN, DeepAR, DSSM, and LSTNet.

**3.2.1 ARIMA**

ARIMA model has been created using the PMDARIMA library and hyper-parameters such as the number of auto-regressive terms (p), number of non-seasonal differences needed for stationarity (d), and number of lagged forecast errors in the prediction equation (q). The hyper parameters passed are shown below.

- p – 0, 1, 2
- q – 0, 1, 2
- Test to determine 'd' – Augmented Dickey-Fuller Test

The best model hyperparameters for ARIMA are (0,1,1).

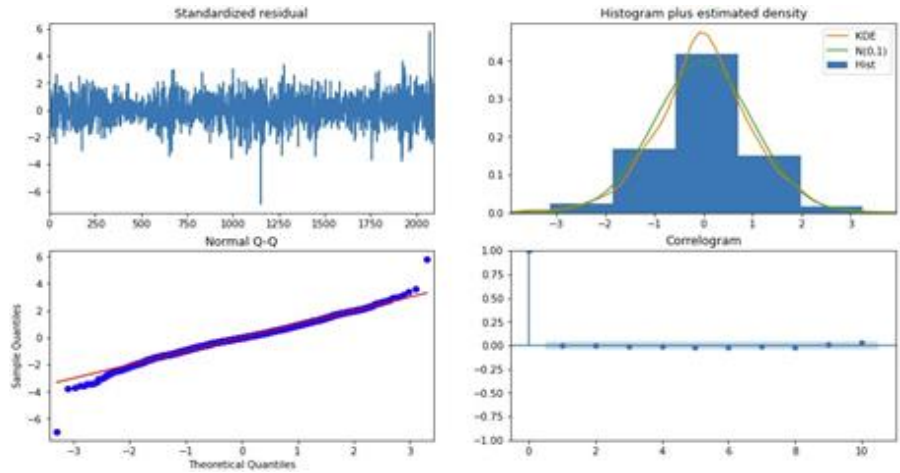

*Figure 3. ARIMA*

### 3.2.2 SARIMA

The model architecture is like ARIMA, except that SARIMA has an additional seasonal component. In addition to the existing hyperparameters of p, q, and d, seasonal hyperparameters, as mentioned below, are passed into the model. The best model hyperparameters for SARIMA are (0,1,1) (0,0,1) [30], as discussed below.

- P – Seasonal Autoregressive order = 0
- D – Seasonal Difference order = 0
- Q – Seasonal Moving Average order = 1
- m – Number of timesteps for a single seasonal period

### 3.2.3 SARIMAX

SARIMAX is similar to SARIMA except for the addition of an exogenous variable which is used as an additional feature in the learning process.

### 3.2.4 Deep Factor RNN (DF-RNN)

The Deep Factor RNN is a model that incorporates two significant factors, namely global and local fluctuations, to govern the progression of a given time series. These factors are learned using separate RNN models, each controlled by specific hyperparameters that determine the model's architecture, as illustrated in Table 1.

Table 1. Global vs. Local RNN Hyperparameters of DF-RNN

| Hyper-parameters | Global RNN model | Local RNN model |
|---|---|---|
| Number of hidden layers | 1 | 5 |
| Number of neurons in the hidden layer | 50 | 5 |
| Type of Cell | LSTM | LSTM |

### 3.2.5 Deep AR

Deep AR, a probabilistic forecasting model, utilizes a global model to learn jointly from multiple related time series using negative binomial likelihood. The model is constructed on a recurrent neural network based on Long Short-Term Memory (LSTM) architecture. Hyperparameters, depicted in Table 2, are used in model building.

Table 2. Deep AR hyperparameters

| Hyper-parameters | Deep AR model |
|---|---|
| Number of LSTM layers | 3 |
| Number of LSTM Cells | 40 |
| Scaling | Enabled |
| Learning rate | 0.001 |

### 3.2.6 Deep State Space Model

The Deep State Space model [27] works on the principles of parametrizing linear state space of individual time series with a jointly learned recurrent neural network.

### 3.2.7 DeepRenewal

Deep Renewal Processes are a probabilistic intermittent demand forecasting method. This method builds upon Croston's framework and molds its variants into a renewal process. The random variables, $M$ (The demand size at non-zero demand point) and $Q$ (The inter-demand interval) are estimated using a separate RNN.

Table 3. DeepRenewal hyperparameters

| Hyper-parameters | Deep Renewal |
|---|---|
| Skip size for skip RNN layer | 3 |
| Auto regressive window size – Linear | 40 |
| Learning Rate | 0.001 |

# 3 Performance Measurement

Performance measurement of the forecasting model has been done using evaluation metrics such as MSE, RMSE, MAE, MAPE and custom metrics such as POCID and Theil's U.

### 4.1 Mean Square Error /Root Mean Square Error

Mean squared error is the mean of the squared error between the target variable (original observation) and the output variable (the predicted variable) in a given time series. Root mean squared error applies a square root to the MSE. The mathematical representation is shown in Equation 1.

$$\text{MSE} = 1/N \sum_{i=1}^{N} (\text{target}_i - \text{output}_i)^2$$

**Equation 1.** MSE

### 4.2 Mean Absolute Percentage Error

Mean Absolute percentage error defines the percentage difference between the target variable and the output variable w.r.t the output variable. The mathematical representation is shown in Equation 2. Lower the value of MAPE better the model performance.

$$\text{MAPE} = 100/N \sum_{i=1}^{N} (\text{target}_i - \text{output}_i)/\text{output}_i$$

**Equation 2.** MAPE

### 4.3 Mean Absolute Error

Mean absolute error is the difference between the target and output variables. The lower the MAE better is the model performance. The mathematical representation is shown in Equation 3

$$\text{MSE} = 1/N \sum_{i=1}^{N} (\text{target}_i - \text{output}_i)$$

**Equation 3.** MAE

### 4.4 POCID – Prediction on Change in Direction

The prediction of change in direction is the percentage of the number of correct decisions in predicting whether the time series in the next time interval will increase or decrease. The mathematical representation is shown in Equation 4. The higher the value of POCID, the better the model performance.

$$POCID = 100 * \frac{\sum_{t=1}^{N} D_t}{N}$$

$$\text{where } D_t = \begin{cases} 1, & \text{if } (target_t - target_{t+1})(output_t - output_{t+1}) > 0 \\ 0, & \text{otherwise} \end{cases}$$

**Equation 4.** POCID

### 4.5 Theil's U

$$\text{THEIL's U (Normalized mean square error)} = \frac{\sum_{t=1}^{N}(target_t - output_t)^2}{\sum_{t=1}^{N}(output_t - output_{t-1})^2}$$

**Equation 5.** Theil's U

Theil's U is similar to the mean squared error except that the error is normalized w.r.t output variable of the previous time interval. U lesser than 1 indicates a better performance of the model. A value equal to 1 indicates a random model and a value greater than 1 indicates a model worse than a random model. Hence it is ideal for achieving a U of value 0. The mathematical representation is shown in Equation 5.

## 4 Results

As explained in the preceding sections, models have been evaluated using the metrics mentioned above on both machine learning and deep learning models.

### 5.1 Forecasting using ARIMA, SARIMA, and SARIMAX

Forecasting has been done over a 36-day horizon. From Table 4, it is evident that there is no significant difference in the model performance among the three baseline models. Furthermore, on analysis of standardized residuals for each forecasted point by all three models, they are uniformly distributed around the mean of zero.

**Table 4.** Results

| Models  | MSE         | MAE       | RMSE      | MAPE     |
|---------|-------------|-----------|-----------|----------|
| ARIMA   | 3003730.628 | 1387.5576 | 1733.1274 | 0.118263 |
| SARIMA  | 3009965     | 1391.598  | 1734.925  | 0.118803 |
| SARIMAX | 3002241     | 1387.658  | 1732.698  | 0.118351 |

## 5.2 Forecasting using Facebook and Prophet

The performance of the Facebook Prophet was assessed using cross-validation on the provided time series for various forecast horizons between 36 and 364 days. It was noted that the MSE and MAPE values increase linearly as the forecast horizon increases. This suggests that an increase in forecast horizon leads to a rise in prediction error. Moreover, when compared to the baseline models, namely ARIMA, SARIMA and SARIMAX, Prophet exhibited a relatively lower performance in terms of RMSE and MAPE.

Table 5. ARIMA, SARIMA, SARIMAX Evaluation Metrics

| Error | 36 days | 364 days | Average |
|---|---|---|---|
| MSE | 203,144 | 5,000,000 | 2,522,746 |
| RMSE | Not Done | Not Done | 1588.31 |
| MAPE | 0.05 | 0.19 | **0.12** |
| POCID | Not Done | Not Done | **80.48** |
| Theil's U | Not Done | Not Done | 494.96 |

## 5.3 Forecasting using Deep Learning Models on 100% Train Data

The table presents the performance evaluation of various deep learning models for time-series forecasting based on five evaluation metrics: MSE, RMSE, POCID, Theil's U, and MAPE. The models include RNN, GRU, LSTM, DF-RNN, DeepAR, DSSM, and Deep Renewal. Among these models, DeepAR performed the best with the lowest values for all evaluation metrics, indicating its high accuracy in forecasting. The LSTM and GRU models also performed well with relatively low values for all metrics. The DF-RNN model performs better than the DSSM and Deep Renewal models but not as well as the others. Therefore, the ranking of the models based on better performance is DeepAR, GRU, LSTM, DF-RNN, RNN, DSSM, and Deep Renewal.

Table 6. Deep Learning Models Performance on 100% train data

| Model | MSE | RMSE | POCID | Thelis'U | MAPE |
|---|---|---|---|---|---|
| RNN | 21341000 | 4619.6 | 49 | 30790 | 36.1 |
| GRU | 53167 | 230.6 | 52 | 892 | 5.5 |
| LSTM | 171396 | 414 | 52 | 898 | 17.4 |
| DF-RNN | 3754898 | 1937.8 | 25 | 1168 | 0.12 |
| DeepAR | **35600** | **188.7** | 75 | **12** | **0.01** |
| DSSM | 25041182 | 5004 | 75 | 7767 | 0.29 |
| Deep Renewal | 246363673 | 15696 | 75 | 77148 | 1.0 |

### 5.4 Forecasting using Deep Learning Models on 50% Train Data

Based on the metrics in the table, the DeepAR model seems to be the best performer, followed by the LSTM, GRU, and RNN models. The DF-RNN, DSSM, and Deep Renewal models performed poorly than the others.

Table 7. Deep Learning Models Performance on 50% train data

| Model | MSE | RMSE | POCID | Thelis'U | MAPE |
|---|---|---|---|---|---|
| RNN | 1779572 | 1334 | 52 | 883 | 16.5 |
| GRU | 330895 | 575 | 52 | 910 | 17.3 |
| LSTM | 438402 | 662 | 51 | 889 | 17.1 |
| DF-RNN | 56715481 | 7531 | 75 | 1168 | 0.12 |
| DeepAR | **2823** | **53.1** | 50 | **0.83** | **0.003** |
| DSSM | 303172659 | 17411 | 75 | 94228 | 0.99 |
| Deep Renewal | 245658060 | 15673 | 50 | 77269 | 0.99 |

### 5.5 Forecasting using Deep Learning Models on 25% Train Data

The table shows that the DeepAR model performs best with the lowest MSE, RMSE, Theil's U, and MAPE values, even with 25% of the actual train data. GRU and LSTM models perform similarly with slightly higher values of the evaluation metrics.

Table 8. Deep Learning Models Performance on 25% train data

| Model | MSE | RMSE | POCID | Thelis'U | MAPE |
|---|---|---|---|---|---|
| RNN | 3970916 | 1993 | 52 | 1114 | 18.2 |
| GRU | 252760 | 503 | 53 | 893 | 17.3 |
| LSTM | 2718187 | 1649 | 52 | 848 | 17.2 |
| DF-RNN | 56715481 | 4985 | 25 | 7719 | 0.32 |
| DeepAR | **1886** | **43** | **50** | **0.76** | **0.002** |
| DSSM | 2442782 | 1563 | 50 | 760 | 0.09 |
| Deep Renewal | 245151974 | 15657 | 50 | 77243 | 0.99 |

Overall, Deep AR and GRU consistently performed the best across all tables, regardless of the percentage of training data used. Surprisingly these models' performance was consistent despite lowering the train data.

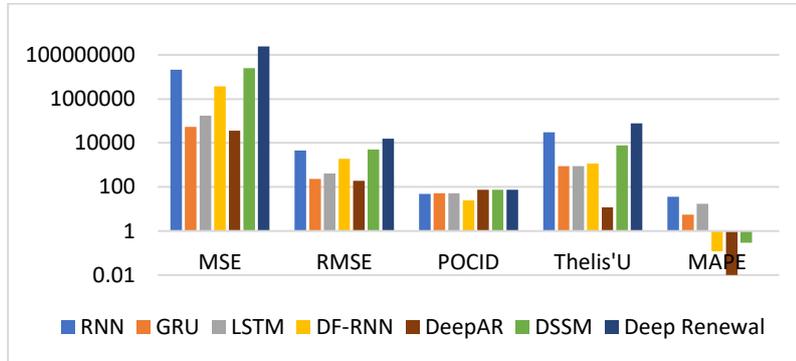

*Figure 4. Evaluation metrics of all models on 100% train data*

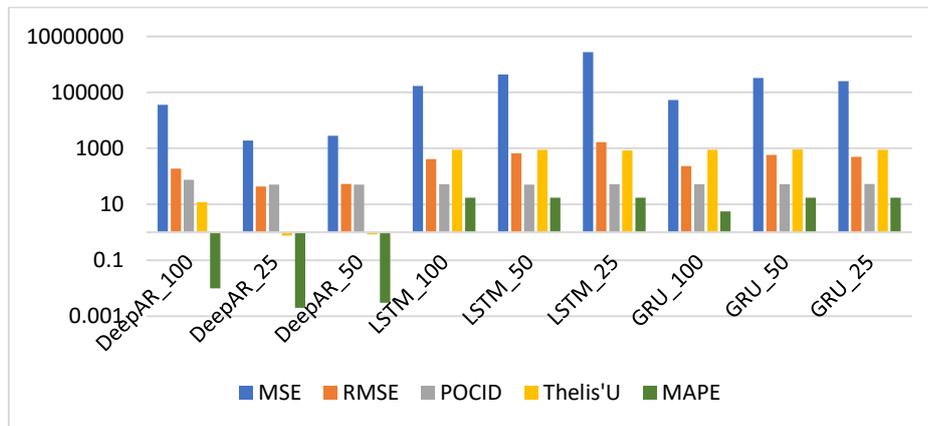

*Figure 5. Evaluation metrics of best three models on varying training data sizes*

## 5  Conclusion

This work compares traditional machine learning models with cutting-edge deep learning architectures for time series forecasting on stock market indices. The study employs several metrics, including MAE, MSE, RMSE, MAPE, POCID, and Theil's U, to evaluate the performance of the models. The results of the experiments demonstrate that state-of-the-art deep neural networks such as DeepAR and GRU outperform traditional forecasting models such as ARIMA, SARIMA, and SARIMAX. Moreover, DeepAR is stable across varying training data sizes and is consistent on all metrics. Furthermore, the study highlights the superiority of recurrent neural networks, their variants, such as LSTMs, for handling stock indices datasets, and their ability to outperform conventional machine learning and statistical-based algorithms. This makes them suitable for deployment in real-world scenarios. However, the study is limited by the use of univariate Stock market datasets. Future research on multivariate datasets could be explored to further establish the superiority of deep learning networks in time series forecasting.

## References

[1]  T. C. Nokeri, "Forecasting Using ARIMA, SARIMA, and the Additive Model," in *Implementing Machine Learning for Finance: A Systematic Approach to Predictive Risk and Performance Analysis for Investment Portfolios*, T. C. Nokeri, Ed., Berkeley, CA: Apress, 2021, pp. 21–50. doi: 10.1007/978-1-4842-7110-0_2.

[2]  S. Wang, C. Li, and A. Lim, "Why Are the ARIMA and SARIMA not Sufficient." arXiv, Mar. 02, 2021. Accessed: Jun. 06, 2023. [Online]. Available: http://arxiv.org/abs/1904.07632

[3]  U. M. Sirisha, M. C. Belavagi, and G. Attigeri, "Profit Prediction Using ARIMA, SARIMA and LSTM Models in Time Series Forecasting: A Comparison," *IEEE Access*, vol. 10, pp. 124715–124727, 2022, doi: 10.1109/ACCESS.2022.3224938.

[4]  J. Sen and S. Mehtab, "Long-and-Short-Term Memory (LSTM) NetworksArchitectures and Applications in Stock Price Prediction," in *Emerging Computing Paradigms*, John Wiley & Sons, Ltd, 2022, pp. 143–160. doi: 10.1002/9781119813439.ch8.

[5]  Y. Wang, A. Smola, D. Maddix, J. Gasthaus, D. Foster, and T. Januschowski, "Deep Factors for Forecasting," in *Proceedings of the 36th International Conference on Machine Learning*, PMLR, May 2019, pp. 6607–6617. Accessed: Jun. 09, 2023. [Online]. Available: https://proceedings.mlr.press/v97/wang19k.html

[6]  S. S. Rangapuram, M. W. Seeger, J. Gasthaus, L. Stella, Y. Wang, and T. Januschowski, "Deep state space models for time series forecasting," in *Advances in Neural Information Processing Systems*, 2018, pp. 7785–7794.

[7]  D. Salinas, V. Flunkert, J. Gasthaus, and T. Januschowski, "DeepAR: Probabilistic forecasting with autoregressive recurrent networks," *International Journal of Forecasting*, vol. 36, no. 3, pp. 1181–1191, Jul. 2020, doi: 10.1016/j.ijforecast.2019.07.001.

[8]  R. D. Snyder, J. K. Ord, and A. Beaumont, "Forecasting the intermittent demand for slow-moving inventories: A modelling approach," *International Journal of Forecasting*, vol. 28, no. 2, pp. 485–496, 2012.

[9]  J. G. De Gooijer and R. J. Hyndman, "25 years of time series forecasting," *International Journal of Forecasting*, vol. 22, no. 3, pp. 443–473, Jan. 2006, doi: 10.1016/j.ijforecast.2006.01.001.

[10] P. Newbold, "The Principles of the Box-Jenkins Approach," *Operational Research Quarterly (1970-1977)*, vol. 26, no. 2, pp. 397–412, 1975, doi: 10.2307/3007750.

[11] D. F. Findley, B. C. Monsell, W. R. Bell, M. C. Otto, and B.-C. Chen, "New Capabilities and Methods of the X-12-ARIMA Seasonal-Adjustment Program," *Journal of Business & Economic Statistics*, vol. 16, no. 2, p. 127, Apr. 1998, doi: 10.2307/1392565.

[12] A. W. Lo, H. Mamaysky, and J. Wang, "Foundations of Technical Analysis: Computational Algorithms, Statistical Inference, and Empirical Implementation." Rochester, NY, Mar. 01, 2000. Accessed: Jun. 21, 2022. [Online]. Available: https://papers.ssrn.com/abstract=228099

[13] R. Webby and M. O'Connor, "Judgemental and statistical time series forecasting: a review of the literature," *International Journal of Forecasting*, vol. 12, no. 1, pp. 91–118, Mar. 1996, doi: 10.1016/0169-2070(95)00644-3.

[14] O. B. Sezer, M. U. Gudelek, and A. M. Ozbayoglu, "Financial time series forecasting with deep learning : A systematic literature review: 2005–2019," *Applied Soft Computing*, vol. 90, p. 106181, May 2020, doi: 10.1016/j.asoc.2020.106181.

[15] G. Lai, W.-C. Chang, Y. Yang, and H. Liu, "Modeling Long- and Short-Term Temporal Patterns with Deep Neural Networks." arXiv, Apr. 18, 2018. Accessed: Jun. 21, 2022. [Online]. Available: http://arxiv.org/abs/1703.07015

[16] D. Salinas, M. Bohlke-Schneider, L. Callot, R. Medico, and J. Gasthaus, "High-dimensional multivariate forecasting with low-rank Gaussian Copula Processes," in *Advances in Neural Information Processing Systems*, Curran Associates, Inc., 2019. Accessed: Jun. 21, 2022. [Online]. Available: https://proceedings.neurips.cc/paper/2019/hash/0b105cf1504c4e241fcc6d519ea962fb-Abstract.html

[17] B. N. Oreshkin, D. Carpov, N. Chapados, and Y. Bengio, "N-BEATS: Neural basis expansion analysis for interpretable time series forecasting." arXiv, Feb. 20, 2020. Accessed: Jun. 21, 2022. [Online]. Available: http://arxiv.org/abs/1905.10437

[18] K. Khare, O. Darekar, P. Gupta, and V. Z. Attar, "Short term stock price prediction using deep learning," in *2017 2nd IEEE International Conference on Recent Trends in Electronics, Information & Communication Technology (RTEICT)*, May 2017, pp. 482–486. doi: 10.1109/RTEICT.2017.8256643.

[19] C. Guo, X. Kang, J. Xiong, and J. Wu, "A New Time Series Forecasting Model Based on Complete Ensemble Empirical Mode Decomposition with Adaptive Noise and Temporal Convolutional Network," *Neural Process Lett*, Oct. 2022, doi: 10.1007/s11063-022-11046-7.

[20] J. Cao, Z. Li, and J. Li, "Financial time series forecasting model based on CEEMDAN and LSTM," *Physica A: Statistical Mechanics and its Applications*, vol. 519, pp. 127–139, Apr. 2019, doi: 10.1016/j.physa.2018.11.061.

[21] F. Zhou, H. Zhou, Z. Yang, and L. Gu, "IF2CNN: Towards non-stationary time series feature extraction by integrating iterative filtering and convolutional neural networks," *Expert Systems with Applications*, vol. 170, p. 114527, May 2021, doi: 10.1016/j.eswa.2020.114527.

[22] S. Atha and B. K. Bolla, "Do Deep Learning Models and News Headlines Outperform Conventional Prediction Techniques on Forex Data?," in *Advances in Distributed Computing and Machine Learning*, R. R. Rout, S. K. Ghosh, P. K. Jana, A. K. Tripathy, J. P. Sahoo, and K.-C. Li, Eds., in Lecture Notes in Networks and Systems. Singapore: Springer Nature, 2022, pp. 413–423. doi: 10.1007/978-981-19-1018-0_35.

[23] R. Wen, K. Torkkola, B. Narayanaswamy, and D. Madeka, "A Multi-Horizon Quantile Recurrent Forecaster." arXiv, Jun. 28, 2018. doi: 10.48550/arXiv.1711.11053.

[24] S. S. Makridakis Evangelos; Assimakopoulos, Vassilios, "The M4 Competition: Results, findings, conclusion and way forward," *International Journal of Forecasting*, vol. 34, no. 4, pp. 802–808, 2018, doi: 10.1016/j.ijforecast.2018.06.001.

[25] T. G. Dietterich, "Ensemble Methods in Machine Learning," in *Multiple Classifier Systems*, in Lecture Notes in Computer Science. Berlin, Heidelberg: Springer, 2000, pp. 1–15. doi: 10.1007/3-540-45014-9_1.

[26] A. Alexandrov *et al.*, "GluonTS: Probabilistic Time Series Models in Python." arXiv, Jun. 14, 2019. Accessed: Jun. 21, 2022. [Online]. Available: http://arxiv.org/abs/1906.05264

[27] S. S. Rangapuram, M. W. Seeger, J. Gasthaus, L. Stella, Y. Wang, and T. Januschowski, "Deep State Space Models for Time Series Forecasting," p. 10.